\documentclass[11pt,letterpaper]{article}
\usepackage{emnlp2017}
\usepackage{times}
\usepackage{latexsym}
\usepackage{graphicx}
\usepackage{hyperref}
\usepackage{amsmath}
\usepackage{CJKutf8}

\emnlpfinalcopy

\def\emnlppaperid{1197}

\title{Learning Chinese Word Representations From Glyphs Of Characters}

\author{Tzu-Ray Su \and Hung-Yi Lee \\
    Dept. of Electrical Engineering, National Taiwan University\\No. 1, Sec. 4, Roosevelt Road, Taipei, Taiwan \\
    {\tt \{b01901007,hungyilee\}@ntu.edu.tw}}

\date{}

\hypersetup{draft}

\begin{document}

\maketitle

\begin{abstract}
  In this paper, we propose new methods to learn Chinese word representations. Chinese characters are composed of graphical components, which carry rich semantics. It is common for a Chinese learner to comprehend the meaning of a word from these graphical components. As a result, we propose models that enhance word representations by character glyphs. The character glyph features are directly learned from the bitmaps of characters by convolutional auto-encoder(convAE), and the glyph features improve Chinese word representations which are already enhanced by character embeddings. Another contribution in this paper is that we created several evaluation datasets in traditional Chinese and made them public.
\end{abstract}

\section{Introduction}
No matter which target language it is, high quality word representations (also known as word ``embeddings") are keys to many natural language processing tasks,  for example, sentence classification \cite{Kim14}, question answering \cite{zhou2015learning}, machine translation \cite{sutskever2014sequence}, etc. Besides, word-level representations are building blocks in producing phrase-level \cite{cho2014learning} and sentence-level \cite{kiros2015skip} representations.

In this paper, we focus on learning Chinese word representations.
A Chinese word is composed of characters which contain rich semantics.
The meaning of a Chinese word is often related to the meaning of its compositional characters. 
Therefore, Chinese word embedding can be enhanced by its compositional character embeddings~\cite{Chen2015,xu2016improve}.
Furthermore, a Chinese character is composed of several graphical components. 
Characters with the same component share similar semantic or pronunciation. 
When a Chinese user encounters a previously unseen character, it is instinctive to guess the meaning (and pronunciation) from its graphical components, so understanding the graphical components and associating them with semantics help people learning Chinese. 
Radicals\footnote{\url{https://en.wikipedia.org/wiki/Radical_(Chinese_characters)}} are the graphical components used to index Chinese characters in a dictionary. By identifying the radical of a character, one obtains a rough meaning of that character, so it is used in learning  Chinese word embedding~\cite{Yin2016} and character embedding~\cite{sun2014radical,li2015component}. 
However, other components in addition to radicals may contain potentially useful information in word representation learning. 

\begin{figure}[h]
\centering
\includegraphics[width=0.45\textwidth]{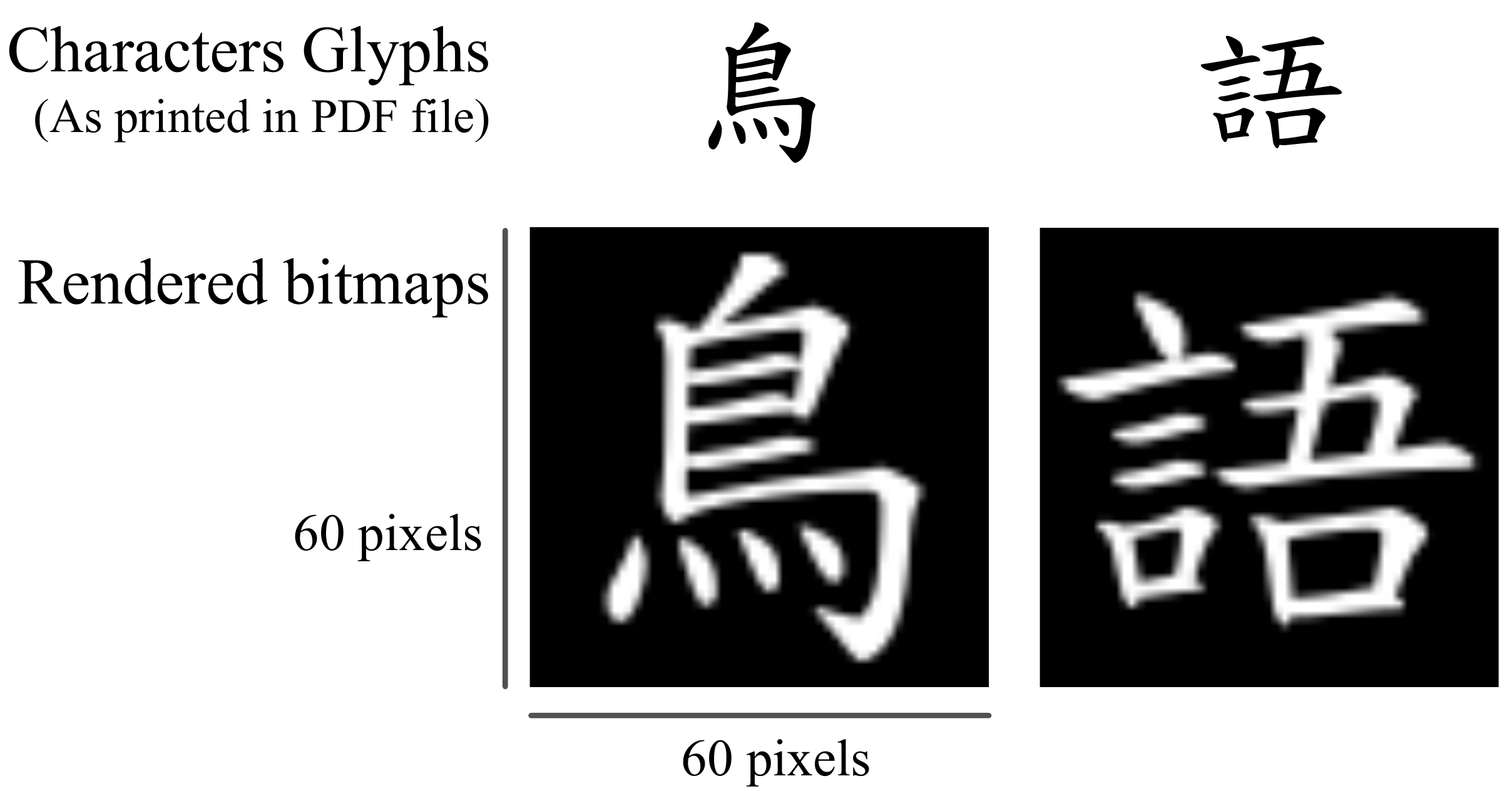}
\caption{A Chinese character is represented as a fixed-size gray-scale image which is referred to as ``character bitmap'' in this paper.}
\label{fig:render_glyph}
\end{figure}

Our research begins with a question: {\em Can machines learn Chinese word representations from glyphs of characters?} 
By exploiting the glyphs of characters as images in word representation learning, all the graphical components in a character are considered, not limited to radicals. 
In our proposed methods, we render character glyphs to fixed-size grayscale images which are referred to as ``character bitmaps'', as illustrated in Fig.\ref{fig:render_glyph}.
A similar idea was also used in \cite{DBLP:journals/corr/LiuLLN17} to help classifying wikipedia article titles into 12 categories.
We use a convAE to extract character features  from the bitmap to represent the glyphs.
It is also possible to represent the glyph of a character by the graphical components in it.
We do not choose this way because there is no unique way to decompose a character, and directly learning representation from bitmaps is more straightforward.
Then we use the models parallel to Skipgram~\cite{mikolov2013efficient} or GloVe~\cite{pennington2014}  to learn word representations from the character glyph features.
Although  we only consider traditional Chinese characters in this paper, and the examples given below are based on the traditional characters, the same ideas and methods can be applied on the simplified characters.

\section{Background Knowledge and Related Works}
To give a clear illustration of our own work, we briefly introduce the representative methods of word representation learning  in Section~\ref{subsec:wordvector_review}.
In Section~\ref{subsec:chinese_review}, we will introduce some of the linguistic properties of Chinese, and then introduce the  methods that utilize these properties to improve word representations.

\subsection{Word Representation Learning} \label{subsec:wordvector_review}
Mainstream research of word representation is built upon the distributional hypothesis, that is, words with similar contexts share similar meanings. Usually a large-scale corpus is used, and word representations are produced from the co-occurrence information of a word and its context. Existing methods of producing word representations could be separated into two families~\cite{levy2015}: count-based family~\cite{turney2010frequency,bullinaria2007extracting}, and prediction-based family. 
Word representations  can be obtained by training a neural-network-based models~\cite{bengio2003neural,collobert2011natural}.
The representative methods are briefly introduced below.

\subsubsection{CBOW and Skipgram}
Both continuous bag-of-words (CBOW) model and Skipgram model train with words and contexts in a sliding local context window~\cite{mikolov2013efficient}. 
Both of them assign each word $w_{i}$ with an embedding $\vec{w}_{i}$.
CBOW predicts the word given its context embeddings, while Skipgram predicts contexts given the word embedding. 
Predicting the occurrence of word/context in CBOW and Skipgram models could be viewed as learning a multi-class classification neural network (the number of classes is the size of vocabulary). In \cite{mikolov2013distributed}, the authors introduced several techniques to improve the performance. Negative sampling is introduced to speed up learning, and subsampling frequent words is introduced to randomly discard training examples with frequent words (such as ``the", ``a", ``of"), and has an effect similar to the removal of stop words.

\subsubsection{GloVe}
Instead of using local context windows, ~\cite{pennington2014} proposed GloVe model. 
Training GloVe word representations begins with creating a co-occurrence matrix $X$ from a corpus, where each matrix entry $X_{ij}$ represents the counts that word $w_{j}$ appears in the context of word $w_{i}$. In \cite{pennington2014}, the authors used a harmonic weighting function for co-occurrence count, that is, word-context pairs with distance $d$ contributes $\frac{1}{d}$ to the global co-occurrence count.

Let $\vec{w}_{i}$ be the word representation of word $w_{i}$, and $\vec{\tilde{w}}_{j}$ be the word representation of word $w_{j}$ as context, GloVe model minimizes the loss: 
$$\sum\limits_{i,j\in \substack{non-zero \\ entries\ of\ X}} f(X_{ij})( \vec{w}_{i}^{T}\vec{\tilde{w}}_{j} + b_{i} + \tilde{b}_{j} -log(X_{ij})),$$
where $b_{i}$ is the bias for word $w_{i}$, and $\tilde{b}_{j}$ is the bias for context $w_{j}$. 
A weighting function $f(X_{ij})$ is introduced because the authors consider rare co-occurrence word-context pairs carry less information than frequent ones, and their contributions to the total loss should be decreased. 
The weighting function $f(X_{ij})$ is defined as below. 
It depends on the co-occurrence count, and the authors set parameters $x_{max}=100$, $\alpha=0.75$.
\[
 f(X_{ij}) = 
  \begin{cases} 
   (\frac{X_{ij}}{x_{max}})^{\alpha} & \text{if } X_{ij} < x_{max} \\
   1       & \text{otherwise }
  \end{cases}
\]

In the GloVe model, each word has 2 representations $\vec{w}$ and $\vec{\tilde{w}}$. The authors suggest using $\vec{w}+\vec{\tilde{w}}$ as the word representation, and reported improvements over using $\vec{w}$ only.

\subsection{Improving Chinese Word Representation Learning} \label{subsec:chinese_review}

\subsubsection{The Chinese Language}

\begin{figure}[h]
\centering
\includegraphics[width=0.5\textwidth]{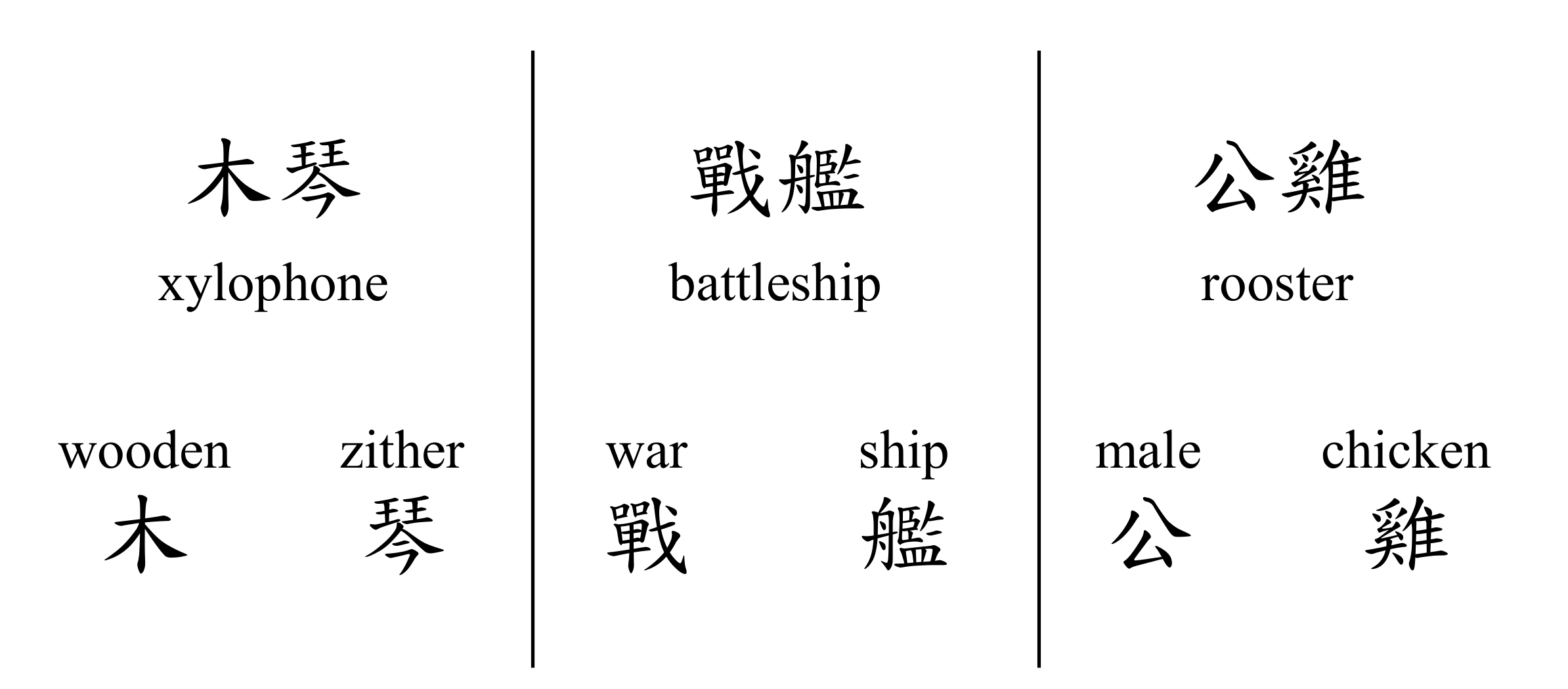}
\caption{Examples of compositional Chinese words. 
Still, the reader should keep in mind that NOT all Chinese words are compositional (related to the meanings of its compositional characters).}
\label{fig:char_compose}
\end{figure}

\begin{CJK}{UTF8}{bsmi}
A Chinese word is composed of a sequence of characters.
The meanings of some Chinese words are related to the composition of the meanings of their characters. 
For example, ``戰艦'' (battleship), is composed of two characters, ``戰'' (war) and ``艦'' (ship). 
More examples are given in Fig.~\ref{fig:char_compose}. 
To  improve Chinese word representations with sub-word information, character-enhanced word embedding (CWE)~\cite{Chen2015} in Section~\ref{subsubsec:CWE} is proposed.
\end{CJK}

\begin{figure}[h]
\centering
\includegraphics[width=0.5\textwidth]{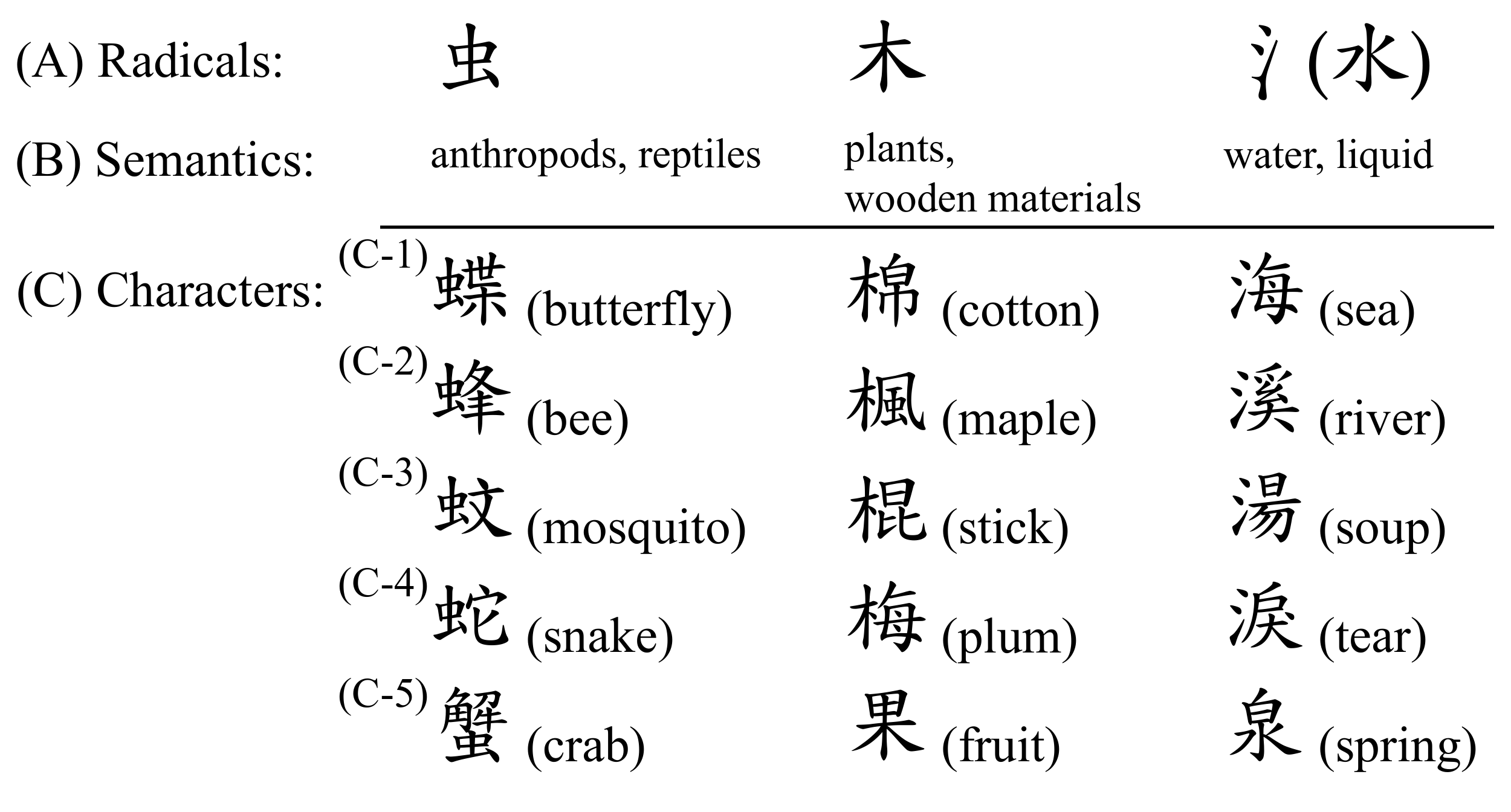}
\caption{Some examples of radicals and the characters containing them.
In rows (C-1) to (C-4), the radicals are at the left hand side of the character, while in row (C-5), the radicals are at the bottom, and may have different of shapes.}
\label{fig:char_clusters}
\end{figure}

A Chinese character is composed of several graphical components. 
Characters with the same component share similar semantic or phonetic properties. 
In a Chinese dictionary characters with similar coarse semantics are grouped into categories for the ease of searching. 
The common graphical component which relates to the common semantic is chosen to index the category, known as a radical. Examples are given in Fig.~\ref{fig:char_clusters}. 
There are  three radicals in  row (A), and their semantic meanings are in row (B).
In each column, there are five characters containing each radical.
It is easy to find that the characters having the same radical have meanings related to the radical in some aspect.
A radical can be put in different positions in a character.
For example, in rows (C-1) to (C-4), the radicals are at the left hand side of a character, but in row (C-5), the radicals are at the bottom.
The shape of a radical can be different in different positions.
For example, the third radical which represents ``water'' or ``liquid'' has different forms when it is at the left hand side or the bottom of a character.
Because radicals serve as a strong semantic indicator of a character, multi-granularity embedding (MGE) \cite{Yin2016} in Section~\ref{subsubsec:MGE} incorporates radical embeddings in learning word representation.

\begin{figure}[h]
\centering
\includegraphics[width=0.45\textwidth]{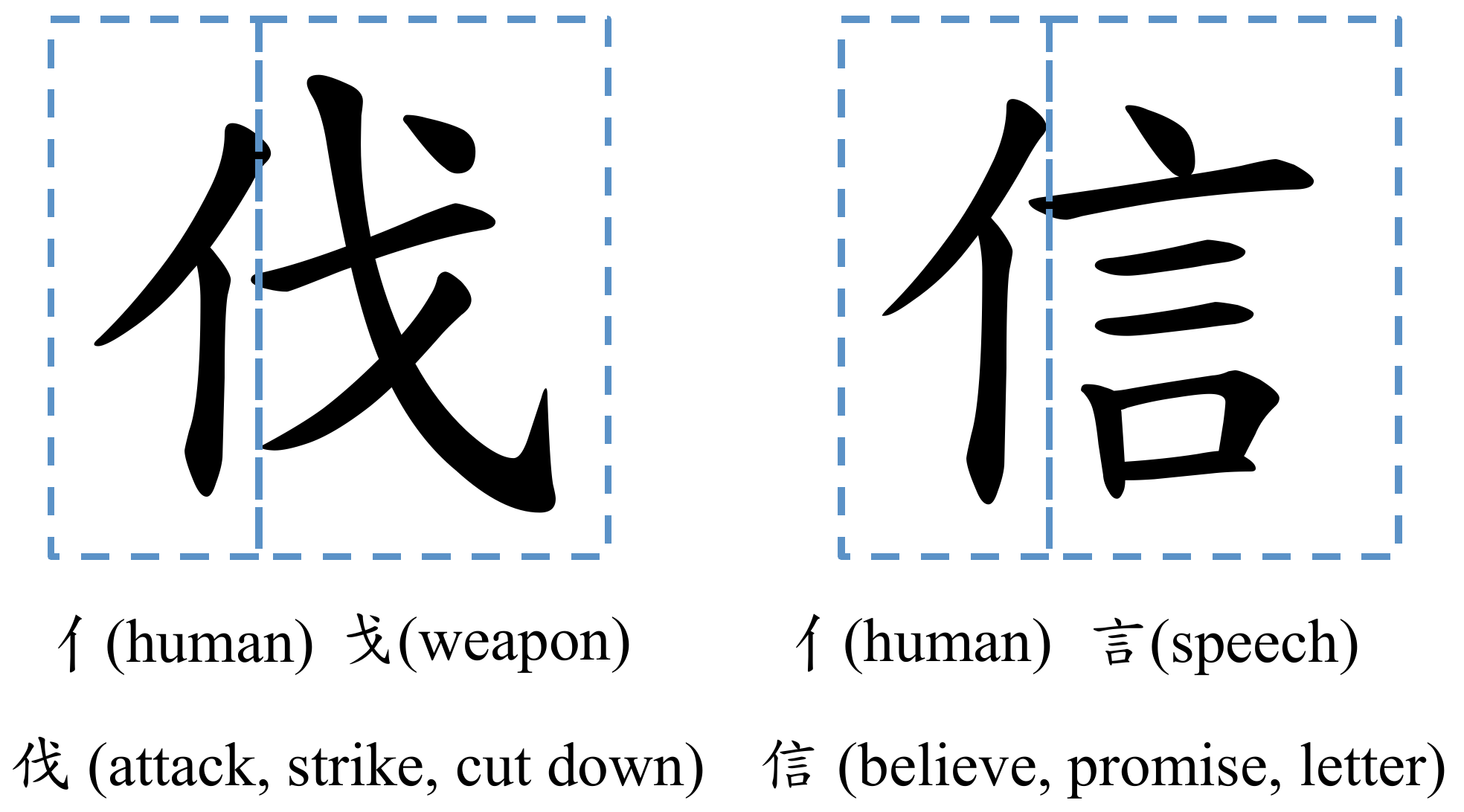}
\caption{\begin{CJK}{UTF8}{gbsn}
Both characters in the figure have the same radical ``亻'' (means humans) at the left hand side, but their meanings are the composition of the graphical components at the right hand side and their radical.
\end{CJK}
}
\label{fig:seman_phone_sim}
\end{figure}

\begin{CJK}{UTF8}{gbsn}
Usually the components other than radicals determine the pronunciation of the characters, but  in some cases they also influence the meaning of a character. 
Two examples are given in Fig.~\ref{fig:seman_phone_sim}\footnote{The two example characters here have the same glyphs in the traditional and simplified Chinese characters.}.
Both characters in Fig.~\ref{fig:seman_phone_sim} have the same radical ``亻'' (means humans) at the left hand side, but the graphical components at the right hand side also have semantic meanings related to the characters.
Considering the left character ``伐''　(means attack).
Its right component ``戈'' means ``weapon'', and the meaning of the character ``伐'' is the composition of the meaning of its two components (a human with a weapon).
None of the previous word embedding approach considers all the components of Chinese characters in our best knowledge.
\end{CJK}

\subsubsection{Character-enhanced Word Embedding (CWE)}\label{subsubsec:CWE}

\begin{figure}[h]
\centering
\includegraphics[width=0.45\textwidth]{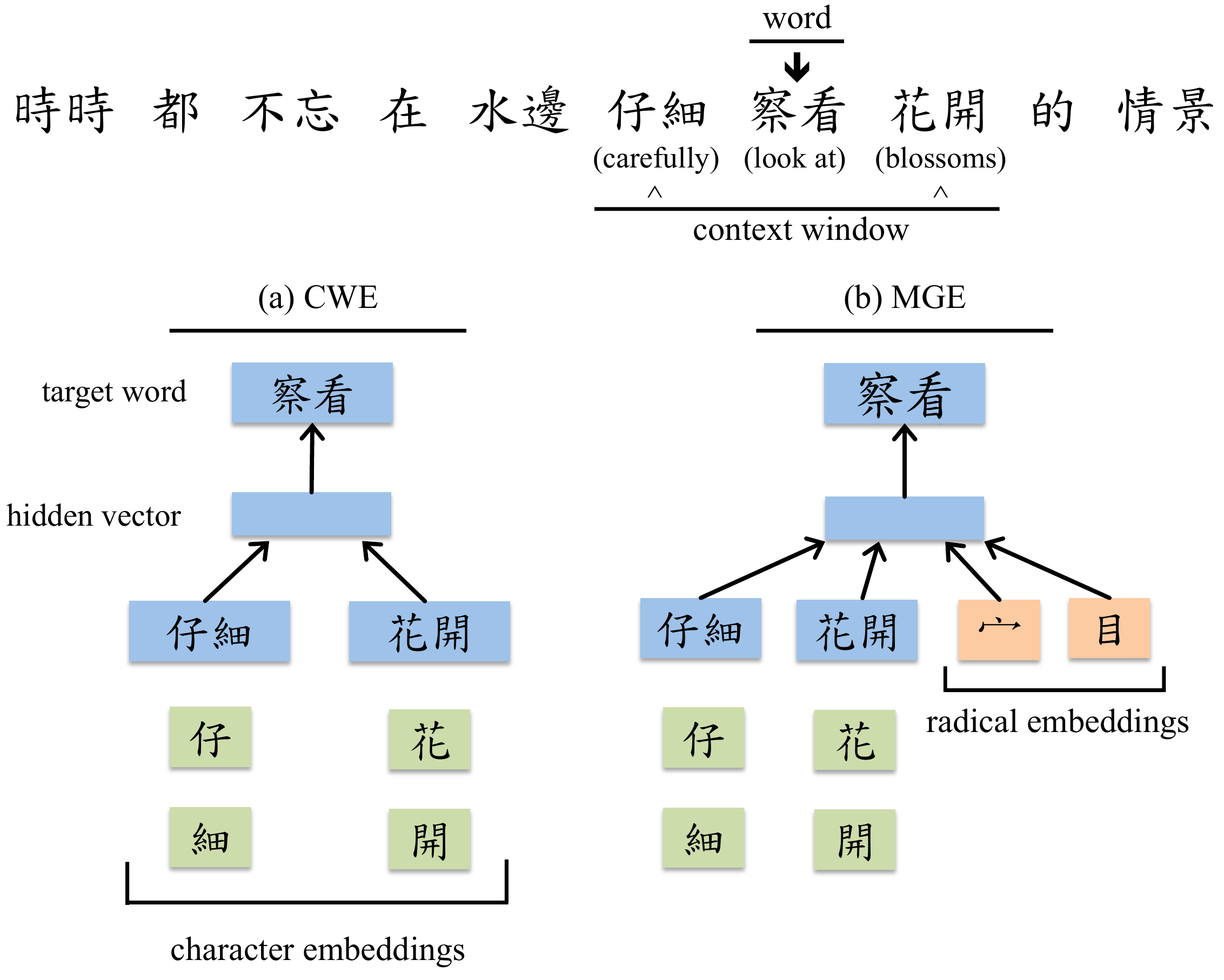}
\caption{Model comparison of Character-enhanced Word Embedding (CWE) and Multi-granularity Embedding (MGE).}
\label{fig:cbow_cwe_mge}
\end{figure}

The main idea of CWE is that word embedding is enhanced by its compositional character embeddings. 
CWE predicts the word from both word and character embeddings of contexts, as illustrated in Fig.~\ref{fig:cbow_cwe_mge} (a). For word $w_{i}$, the CWE word embedding $\vec{w}^{cwe}_{i}$ has the following form:
$$ \vec{w}^{cwe}_{i} = \vec{w}_{i} + \frac{1}{|C(i)|}\sum_{c_{j}\in C(i)}\vec{c}_{j}$$
where $\vec{w}_{i}$ is the word embedding, $\vec{c}_{j}$ is the embedding of the j-th character in $w_{i}$, and $C(i)$ is the set of compositional characters of word $w_{i}$. Mean value of CWE word embeddings of contexts are then used to predict the word $w_{i}$.
 
Sometimes one character has several different meanings, this is known as the ambiguity problem. 
To deal with this, each character is assigned with a bag of embeddings. 
During training, one of the embeddings is picked to form the modified word embedding.
The authors proposed three methods to decide which embedding is picked: position-based, cluster-based, and non-parametric cluster-based character embeddings.

\subsubsection{Multi-granularity Embedding (MGE)}\label{subsubsec:MGE}
Based on CBOW and CWE, ~\cite{Yin2016} proposed MGE, which predicts target word with its radical embeddings and modified word embeddings of context in CWE, as shown in Fig.\ref{fig:cbow_cwe_mge} (b).

There is no ambiguity of radicals, so each radical is assigned with one embedding $\vec{r}$. We denote $\vec{r}_{k}$ as the radical embedding of character $c_{k}$. 
MGE predicts the target word $w_{i}$ with the following hidden vector: 
$$ \vec{h}_{i} =\frac{1}{|C(i)|}\sum_{c_{k}\in C(i)}\vec{r}_{k} + \frac{1}{|W(i)|} \sum_{w_{j}\in W(i)}\vec{w}^{cwe}_{j} $$,
where 
W(i) is the set of contexts words of $w_{i}$, $\vec{w}^{cwe}_{j}$ is the CWE word embedding of $w_{j}$.
MGE picks character embeddings with the position-based method in CWE, and picks radical embeddings
according to a character-radical index built from a dictionary during training.
When non-compositional word is encountered, only the word embedding is used to form $\vec{h}_{i}$. 

\section{Model} \label{sec:model}

We first extract glyph features from bitmaps with the convAE in Section~\ref{subsec:auto}.
The glyph features are used to enhance the existing word representation learning models in Section~\ref{subsec:CWE+glyph}.
In Section~\ref{subsec:RNN+glyph}, we try to learn word representations directly  from the glyph features. 

\subsection{Character Bitmap Feature Extraction} \label{subsec:auto}
\begin{figure}[h]
\centering
\includegraphics[width=0.48\textwidth]{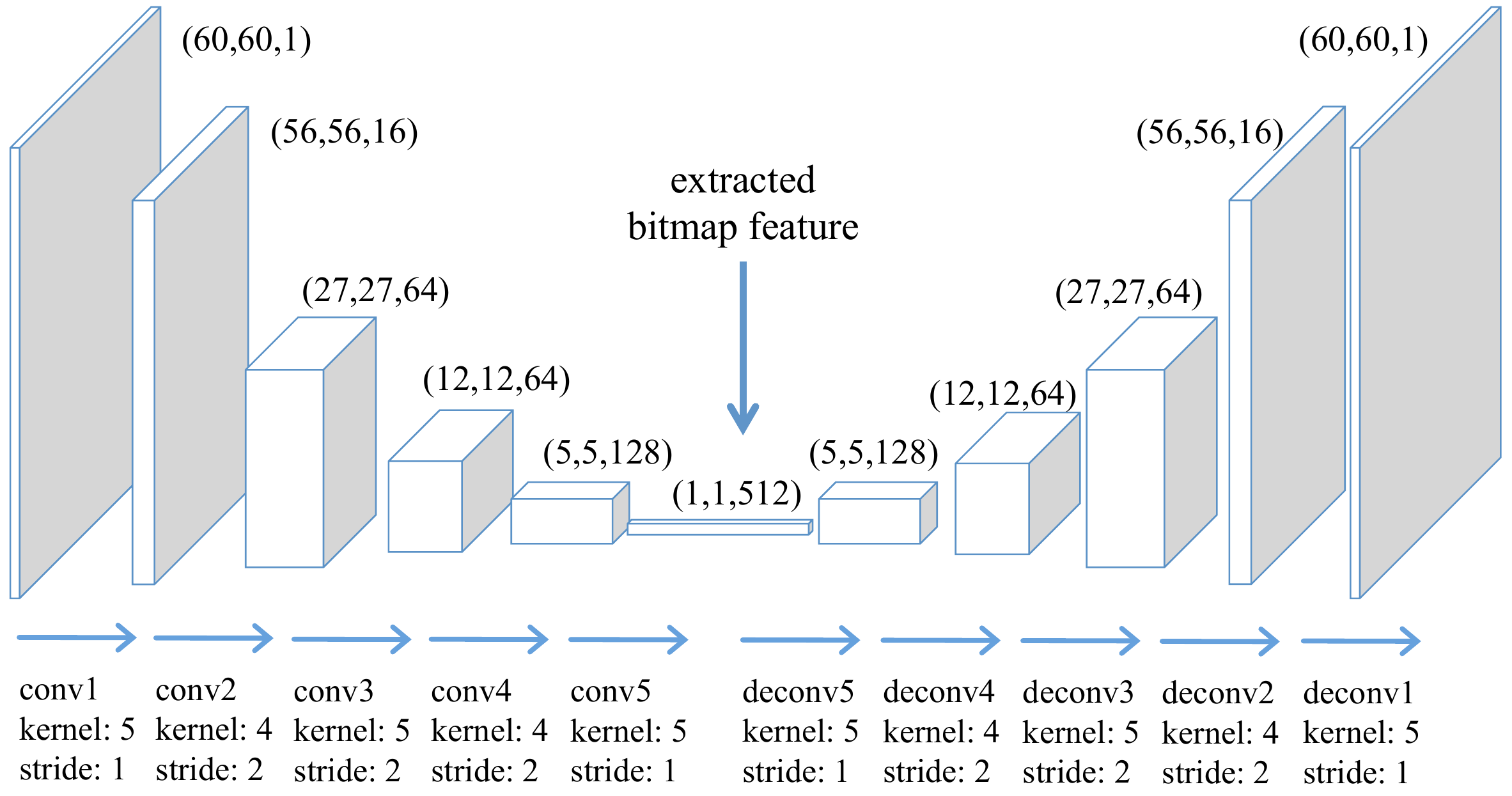}
\caption{The architecture of convAE.}
\label{fig:convae_arch}
\end{figure}

A convAE~\cite{masci2011} is used to reduce the dimensions of rendered character bitmaps and capture high-level features. 
The architecture of the convAE is shown in Fig.~\ref{fig:convae_arch}.
The convAE is composed of 5 convolutional layers in both encoder and decoder.
The stride larger than one is used instead of pooling layers. 
Convolutional and deconvolutional layers on the same level share the same kernel. 
The input image is a 60$\times$60 8-bit grayscale bitmap, and the encoder extracts 512-dimensional  feature. 
The  feature of character $c_{k}$ from the encoder is refer to as character glyph feature $\vec{g}_{k}$ in the paper. 

\subsection{Glyph-Enhanced Word Embedding (GWE)} 
\label{subsec:CWE+glyph}

\subsubsection{Enhanced by Context Word Glyphs}
\label{subsubsec:CWE+P+ctxG}
\begin{figure}[h]
\centering
\includegraphics[width=0.45\textwidth]{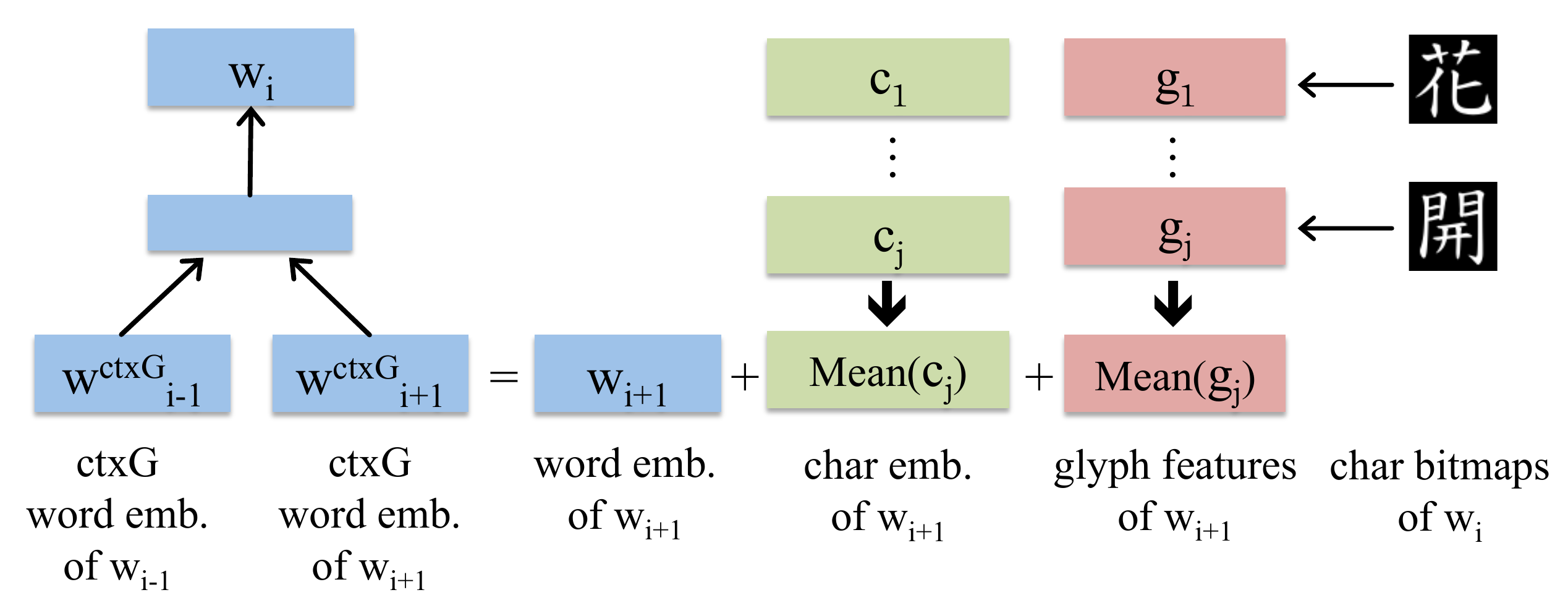}
\caption{Illustration of exploiting context word glyphs. Mean value of character glyph features in the context is added to the hidden vector that predicts target word.}
\label{fig:cwe_p_ctx_glyph}
\end{figure}

We modify CWE model based on CBOW in Section~\ref{subsubsec:CWE} to incorporate context character glyph features (ctxG). 
This modified word embedding $\vec{w}^{ctxG}_{i}$ of word $w_{i}$ has the form: 
$$
\vec{w}^{ctxG}_{i} = \vec{w}_{i} + \frac{1}{|C(i)|}\sum_{c_{j}\in C(i)}(\vec{c}_{j} + \vec{g}_{j}), 
$$
where $C(i)$ is the compositional characters of $w_{i}$ and $\vec{g}_{j}$ is the glyph feature of $c_{j}$. The model predicts target word $w_{i}$ from ctxG word embeddings of contexts, as shown in Fig.\ref{fig:cwe_p_ctx_glyph}.  
The parameters in the convAE are pre-trained, thus not jointly learned with embeddings $\vec{w}$ and $\vec{c}$, so  character glyph features $\vec{g}$ are fixed during training.

\subsubsection{Enhanced by Target Word Glyphs}
\begin{figure}[h]
\centering
\includegraphics[width=0.45\textwidth]{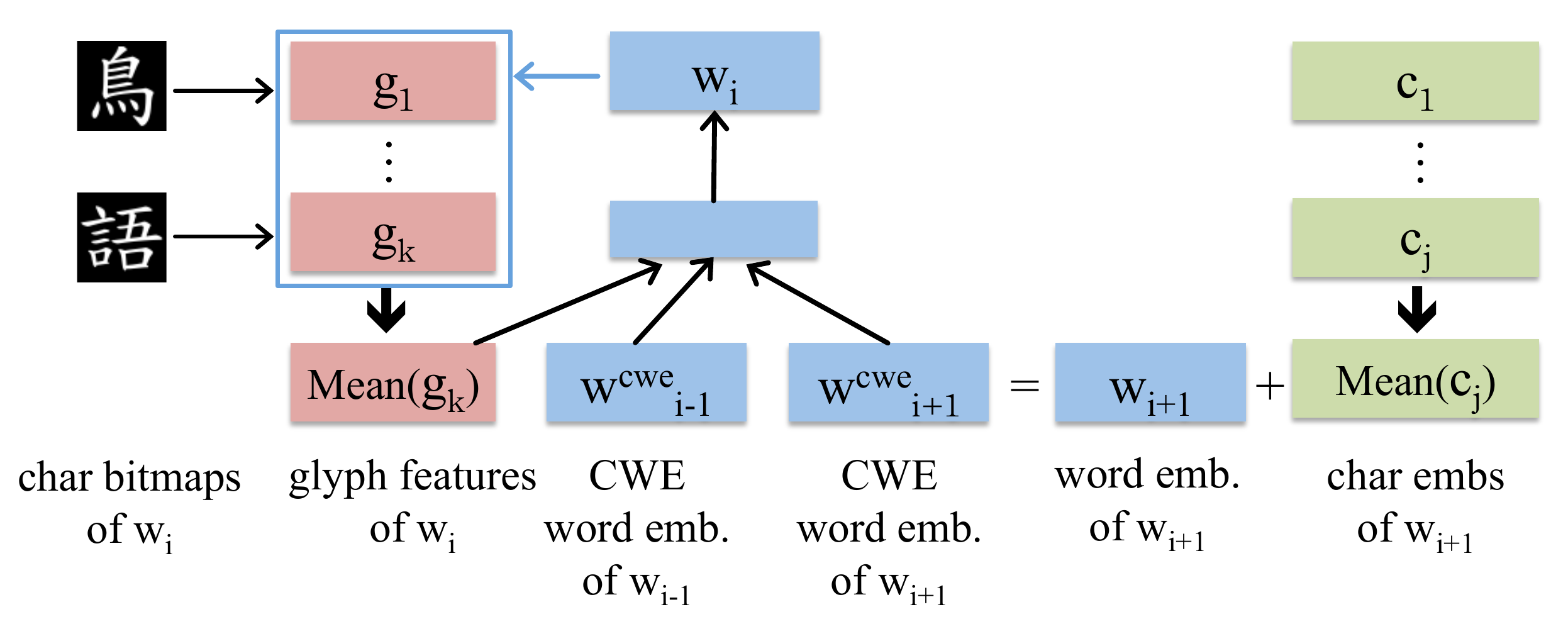}
\caption{Illustration of exploiting target word glyphs. Mean value of character glyph features of target words  help predicting target word itself.}
\label{fig:cwe_p_tar_glyph}
\end{figure}

Here we propose another variant. In this model, the
model structure is the same as in Fig.\ref{fig:cwe_p_ctx_glyph}. The difference lies in the hidden vector used to predict the target word. Instead of adding mean value of character glyph features of the contexts, it adds mean value of glyph feature of the target word (tarG), as shown in Fig.\ref{fig:cwe_p_tar_glyph}. As in Section~\ref{subsubsec:CWE+P+ctxG}, convAE is not jointly learned. 

\subsection{Directly Learn From Character Glyph Features} \label{subsec:RNN+glyph}

\subsubsection{RNN-Skipgram}
\begin{figure}[h]
\centering
\includegraphics[width=0.45\textwidth]{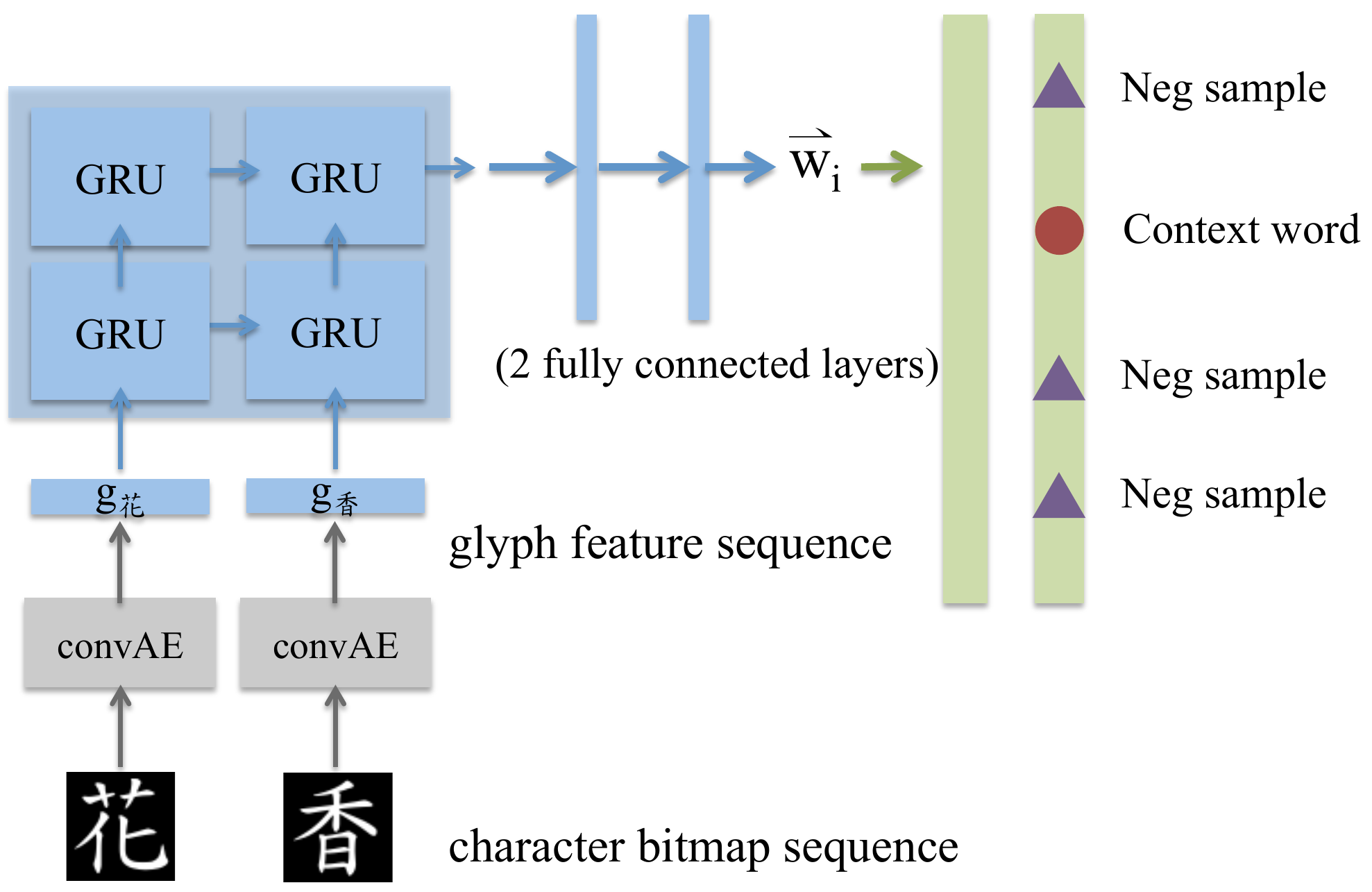}
\caption{Model architecture of RNN-Skipgram model. 
Produced word representation $\vec{w}_{i}$
is used to predict the context of word $w_{i}$.}
\label{fig:rnn_sg}
\end{figure}

We learn word representation $\vec{w}_{i}$ directly from the sequence of character glyph features $\{\vec{g}_{k}, c_{k}\in C(i)\}$ of word $w_{i}$, with the objective of Skipgram. 
As in Fig.\ref{fig:rnn_sg}, a 2-layer Gated Recurrent Units (GRU) \cite{cho2014learning} network followed by 2 fully connected ELU \cite{clevert2015fast} layers produces word representation $\vec{w}_{i}$ from input sequence $\{\vec{g}_{k}\}$ of word $w_{i}$. $\vec{w}_{i}$ is then used to predict the contexts of $w_{i}$.
In the training we use negative sampling and subsampling on frequent words from \cite{mikolov2013distributed}. 

\subsubsection{RNN-GloVe}
\begin{figure}[h]
\centering
\includegraphics[width=0.45\textwidth]{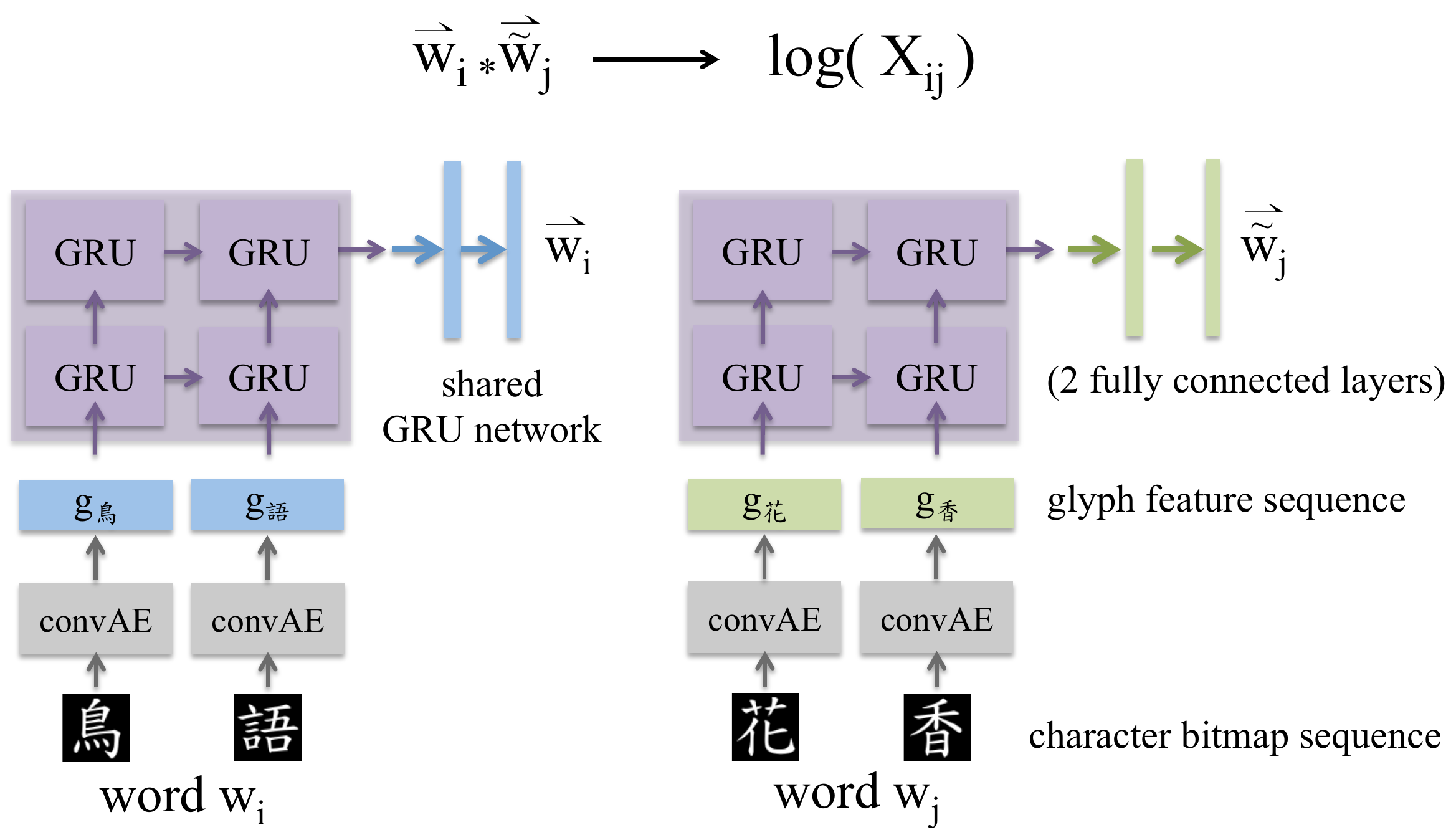}
\caption{Model architecture of RNN-GloVe. A shared GRU network and 2 different sets of fully connected ELU layers produce $\vec{w}_{i}$ and $\vec{\tilde{w}}_{j}$.
Inner product of $\vec{w}_{i}$ and $\vec{\tilde{w}}_{j}$ is the prediction of log co-occurrence $log(X_{ij})$.}
\label{fig:rnn_glove}
\end{figure}

We modify GloVe model to directly learn from character glyph features as in Fig.\ref{fig:rnn_glove}. 
We feed character glyph feature sequence $\{\vec{g}_{k}, c_{k}\in C(i)\}$, $\{\vec{g}_{k^\prime}, c_{k^\prime}\in C(j)\}$ of word $w_{i}$ and context $w_{j}$ to a shared GRU network. 
Outputs of GRU are then fed to two different fully connected ELU layers to produce word representations $\vec{w}_{i}$ and $\vec{\tilde{w}}_{j}$. 
The inner product of $\vec{w}_{i}$ and $\vec{\tilde{w}}_{j}$ is the prediction of log co-occurrence $log(X_{ij})$. We apply the same loss function with weights in GloVe. We follow \cite{pennington2014} and use $\vec{w}_{i}+\vec{\tilde{w}}_{i}$ for evaluations of word representation. \\

\section{Experimental Setup} \label{sec:training_details}

\subsection{Preprocessing}
We learned word representations with traditional Chinese texts from Central News Agency daily newspapers from 1991 to 2002 (Chinese Gigaword, LDC2003T09). 
All foreign words, numerical words, and punctuations were removed. 
Word segmentation was performed using open source python package {\tt jieba}\footnote{\url{https://github.com/fxsjy/jieba}}. In all 316,960,386 segmented words, we extracted 8780 unique characters, and used a true type font (BiauKai) to render each character glyph to a 60$\times$60 8-bit grayscale bitmap. Furthermore, We removed words whose frequency $<=$ 25, leaving 158,565 unique words as the vocabulary set.

\subsection{Extracting Visual Features of Character Bitmap}
Inspired by \cite{zeiler2011adaptive}, layer-wise training was applied to our convAE.
From lower level to higher, the kernel of each layer is trained individually, with other kernels frozen for 100 epochs. Loss function is the Euclidean distance between input and reconstructed bitmap, and we added $l1$ regularization to the activations of convolution layers. 
We chose Adagrad as the optimizing algorithm, and set batch size $=20$ and learning rate $=0.001$.

\begin{figure}[h]
\centering
\includegraphics[width=0.5\textwidth]{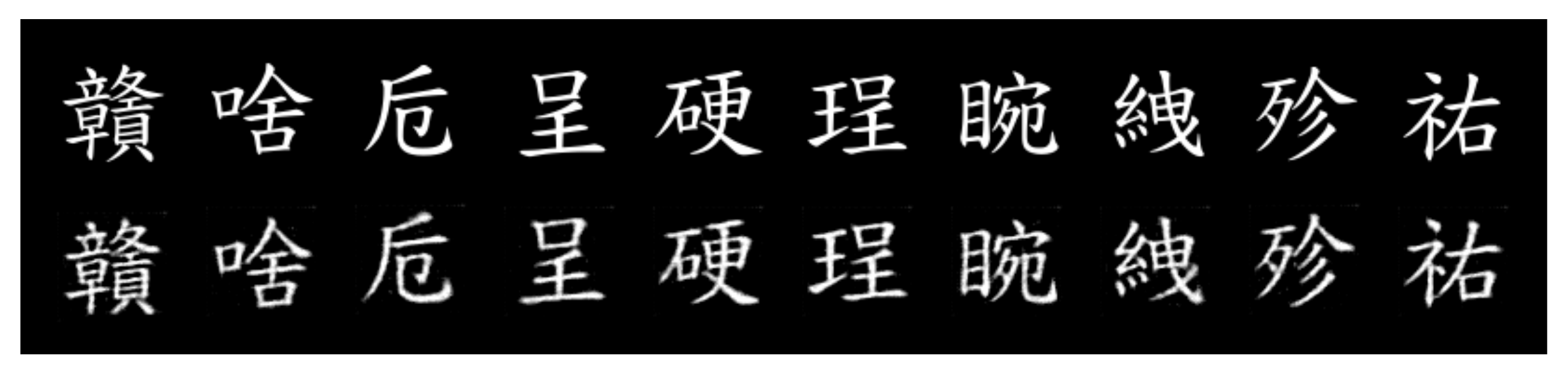}
\caption{The input bitmaps of convAE and their reconstructions.
The input bitmaps are in the upper row, while the reconstructions are in the lower row.}
\label{fig:convae_l5_reconstruct}
\end{figure}

\begin{figure*}[h]
\centering
\includegraphics[width=0.95\textwidth]{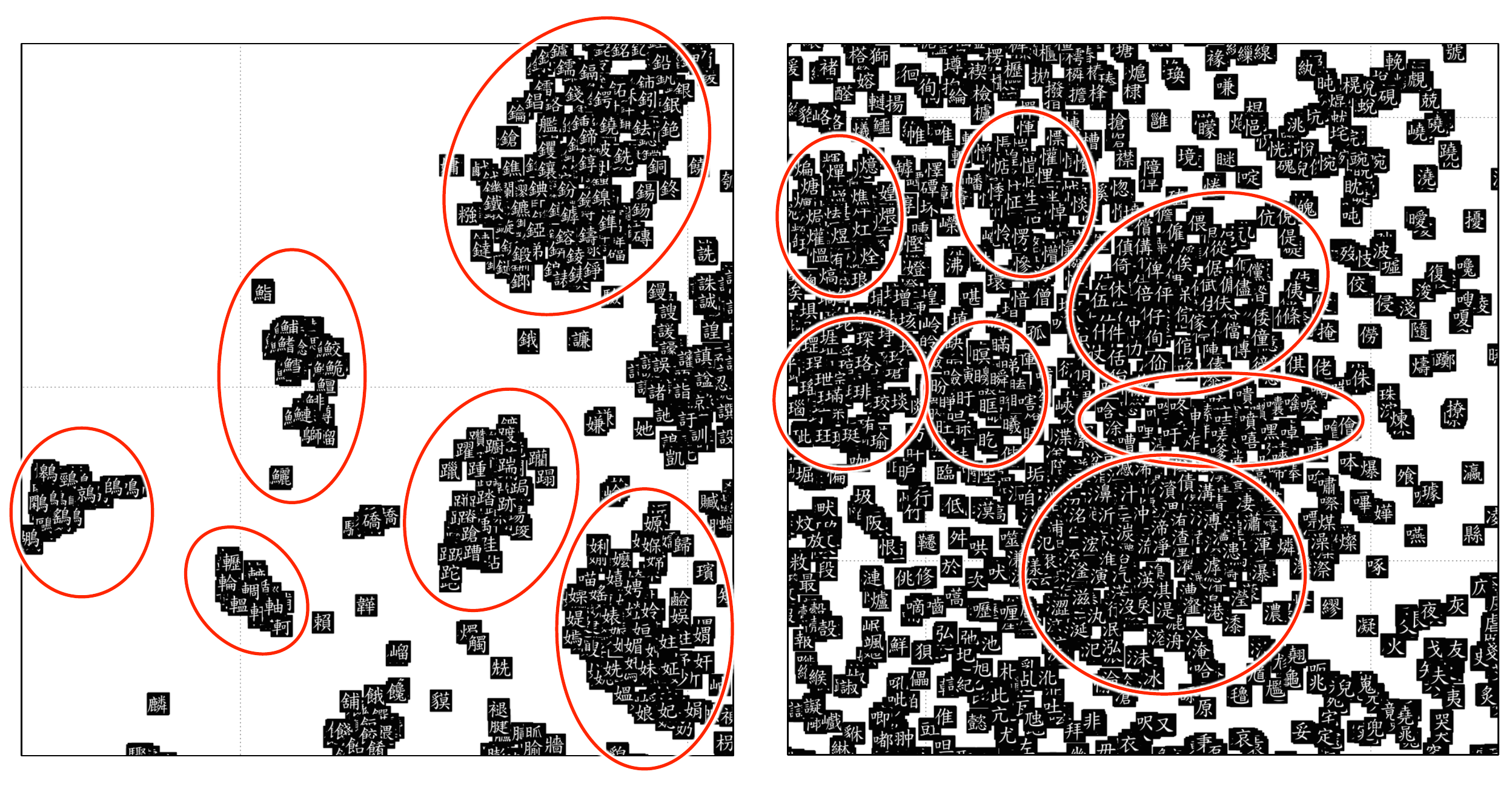}
\caption{Parts of t-SNE visualization of character glyph features. Most of the characters in the ovals share the same components.}
\label{fig:convae_tsne}
\end{figure*}

The comparison between the input bitmaps and their reconstructions is shown in Fig~\ref{fig:convae_l5_reconstruct}.
The input bitmaps are in the upper row, while the reconstructions are in the lower row.
We further visualized the extracted character glyph features with t-SNE~\cite{maaten2008visualizing}.
Part of the visualization result is shown in Fig.~\ref{fig:convae_tsne}.
From Fig.~\ref{fig:convae_tsne}, we found that the  characters with the same components are clustered.
The result shows that the features extracted by the convAE are capable of expressing the graphical information in the bitmaps.

\subsection{Training Details of Word Representations}
We used CWE code\footnote{\url{https://github.com/Leonard-Xu/CWE}} to implement both CBOW and Skipgram, along with the CWE. 
The number of multi-embedding was set to 3.
We modified the CWE code to 
produce GWE representations.
For  CBOW, Skipgram, CWE, GWE and RNN-Skipgram, we used the following hyperparameters.
Context window was set to 5 to both  sides of a word. 
We used 10 negative samples, and threshold $t$ of subsampling was set to $10^{-5}$. 

Since Yin at al. did not publish their code, we followed their paper and reproduced the MGE model. We created the mapping between characters and radicals from the Unihan database\footnote{\url{http://unicode.org/charts/unihan.html}}. Each character corresponds to one of the 214 radicals in this dataset, and the same hyperparameters were used in training as above. Note that we did not separate non-compositional words during training as the original CWE and MGE did.

We used the GloVe code\footnote{\url{https://github.com/stanfordnlp/GloVe}} to train the baseline GloVe vectors.
In construction of co-occurrence matrix for GloVe and RNN-GloVe, we followed the parameter settings of $x_{max}=100$ and $\alpha=0.75$ in ~\cite{pennington2014}. Context window was 5 words to the both sides of a word, and harmonic weighting was  used on co-occurrence counts. 
For the RNN-GloVe   model, we removed entries whose value $< 0.5$ to speed up training. 

RNN-Skipgram and RNN-GloVe generated 200-dimensional word embeddings, while other models generated 512-dimensional word embeddings.

To encourage further research, we published our convAE and embedding models on github\footnote{\url{https://github.com/ray1007/GWE}}. Evaluation datasets were also uploaded, whose details will be explained in Section \ref{sec:eval}.

\section{Evaluation}
\label{sec:eval}

\subsection{Word Similarity}
A word similarity test contains multiple word pairs and their human annotated similarity scores. Word representations are considered good if the calculated similarity and human annotated scores have a high rank correlation.  We computed the Spearman's correlation between human annotated scores and cosine similarity of word representations. 

\begin{CJK*}{UTF8}{bsmi}
Since there is little resource for traditional Chinese, we translated WordSim-240 and WordSim-296 datasets provided by \cite{Chen2015}. Note that this translation is non-trivial. Some frequent words are considered out-of-vocabulary (OOV) due to the different usage between the simplified and traditional. For example, ``butter'' is translated to ``黃油'' in simplified, but ``奶油'' in traditional.
\end{CJK*}
Besides, we manually translated SimLex-999~\cite{hill2016simlex} to traditional Chinese, and used it as the third testing dataset.
We also made these datasets public along with our code.

\begin{table}[h]
\centering
\small
\begin{tabular}{|l|c|c|c|}
\hline
Model & WS-240 & WS-296 & SL-999 \\\hline
\hline
CBOW & \textbf{0.5203} & 0.5550 & 0.3330\\ 
\quad +CWE & 0.4914 & \textbf{0.5553} & 0.3471\\
\quad +CWE+MGE & 0.3767 & 0.2962 & 0.2762\\
\quad +CWE+ctxG & 0.4982 & 0.5549 & \textbf{0.3538}\\
\quad +CWE+tarG & 0.5038 & 0.5503 & 0.3493\\
\hline
Skipgram & \textbf{0.5922} & 0.5876 & 0.3663\\
\quad +CWE & 0.5916 & \textbf{0.5936} & 0.3668\\
\quad +CWE+ctxG & 0.5886 & 0.5856 & \textbf{0.3686}\\
\hline
RNN-Skipgram & 0.3414 & 0.3698 & 0.2464\\
RNN-Glove & 0.2963 & 0.1563 & 0.1010\\
\hline
\end{tabular}
\caption{Spearman's correlation between human annotated scores and cosine similarity of word representations on three datasets: WordSim-240, WordSim-296 and SimLex-999. The higher the values, the better the results.}\label{table:wordsim}
\end{table}

When calculating similarities, word pairs containing OOVs were removed. In Table \ref{table:wordsim}, there are only 237, 284 and 979 word pairs left in WordSim-240, WordSim-296 and SimLex-999, respectively. The results are presented in Table~\ref{table:wordsim}.
The results of ordinary CBOW and Skipgram are shown in the table.
CBOW/Skipgram+CWE represents CWE trained  as CBOW or Skipgram.
For CWE, we only show the results of position-based character embeddings here because the results of cluster-based character embeddings are worse  in the experiments.
We found that CWE only consistently improved the performance on SimLex-999 for both CBOW and Skipgram probably because SimLex-999 contains more words  that could be understood from their compositional characters. 
On SimLex-999, we observed that CWE was better with CBOW than Skipgram. 
We think the reason is that CBOW+CWE predicts the target word with the mean value of all character embeddings in the context, thus has a less noisy feature; however Skipgram+CWE uses character embeddings of an individual word. 
This noisy feature could cause negative effects on predicting the target word. 
The GWEs were learned based on CWE in two ways.
``ctxG'' represents using glyph features of context words, while ``tarG'' represents using glyph features of target words.
The glyph features improved CWE on WordSim-240 and SimLex-999, but not WordSim-296.

As for MGE results, we were not able to reproduce the performance in ~\cite{Yin2016}. We list possible reasons as below: we did not separate non-compositional word during training (character and radical embeddings are not used for these words), and the we created character-radical index from different data source. We conjecture that the first to be the most crucial factor in reproducing MGE. 

The results of  RNN-Skipgram and RNN-GloVe are also in Table~\ref{table:wordsim}.
Their results are not comparable with CBOW and Skipgram. 
From the results, we conclude that it is not easy to produce word representations directly from glyphs. 
We think the reason is that RNN representations are dependent on each other. Updating model parameters for word $w_{i}$ would also change the word representation of word $w_{j}$. As a result it is much more difficult to train such models.

We further inspect the impact of glyph features by doing significance test\footnote{We followed the method described in \url{https://stats.stackexchange.com/questions/17696/}} between proposed methods and existing ones. The p-values of the tests are given in Table \ref{table:wordsim_sigtest}. We found only ``tarG'' method has a p-value less than 0.05 over CWE.
\begin{table}[h]
\centering
\small
\begin{tabular}{|l|c c|}
\hline
& +CWE+ctxG & +CWE+tarG\\\hline
CBOW & 0.085 & 0.215 \\ 
CBOW+CWE & 0.190 & \textbf{0.008} \\
\hline
\end{tabular}
\caption{p-values of significance tests between proposed methods and existing ones.}\label{table:wordsim_sigtest}
\end{table}

\subsection{Word Analogy}
An analogy problem has the following form: {\em ``king'':``queen'' = ``man'':``?''}, and {\em``woman''} is answer to {\em``?''}. By answering the question correctly, the model is considered capable of expressing semantic relationships. Furthermore, the analogy relation could be expressed by vector arithmetic of word representations as shown in ~\cite{mikolov2013distributed}. 
For the above problem, we find word $w_{i}$ such that $w_{i} = \arg\max\limits_{w}\ cos(\vec{w}, \vec{w}_{queen}-\vec{w}_{king}+\vec{w}_{man})$.

As in the previous subsection, we translated the word analogy dataset in \cite{Chen2015} to traditional. 
The dataset contains 3 groups of analogy problems: capitals of countries, (China) states/provinces of cities, and family relations. 
Considering that most capital and city names do not relate to the meaning of their compositional characters, and that we did not separate non-compositional word in our experiments, we proposed a new analogy dataset composed of jobs and places ({\em job\&place}). Nonetheless, there might be multiple corresponding places for a single job. For instance, A {\em``doctor"} could be in a {\em``hospital''} or {\em``clinic''}. In this {\em job\&place} dataset, we provide a set of places for each job. The model is considered to answer correctly as long as the predicted word is in this set.

We take the mean of all word representations of places ($mean(\vec{w}_{places_1})$) for the first job ($job_1$), and find the place for another job ($job_2$) by calculating $w_{i}$ such that $w_{i} = \arg\max\limits_{w}\ cos(\vec{w}, mean(\vec{w}_{places_1})-\vec{w}_{job_1}+\vec{w}_{job_2})$.

\begin{table}[t]
\centering
\small
\begin{tabular}{|l|c|c|c|c|}
\hline
Method & Capital & City & Family & J\&P\\
\hline
\hline
CBOW            & \textbf{0.8006} & \textbf{0.7200} & 0.4228 & 0.3100\\ 
\quad +CWE      & 0.7858 & 0.5829 & 0.4743 & 0.2667\\
\quad +MGE  & 0.0384 & 0.0114 & 0.1287 & 0.0433\\
\quad +CWE+ctxG & 0.7888 & 0.5771 & 0.4963 & 0.2917\\
\quad +CWE+tarG & 0.7858 & 0.5829 & \textbf{0.5184} & 0.2817\\
\hline
Skipgram & \textbf{0.7962} & \textbf{0.8971} & 0.4779 & 0.2317\\
\quad +CWE & 0.7932 & 0.8686 & 0.5404 & 0.2000\\
\quad +CWE+ctxG & 0.7932 & 0.8686 & \textbf{0.5662} & 0.2000\\
\hline
RNN-Skipgram & 0.0000 & 0.0057 & 0.0368 & -\\
RNN-Glove & 0.0281 & 0.0057 & 0.0184 & -\\
\hline
\end{tabular}
\caption{Accuracy of analogy problems for capitals of countries, (China) states/provinces of cities, family relations, and our proposed {\em job\&place} (J\&P) dataset. The higher the values, the better the results.}\label{table:analogy}
\end{table}

The results are shown in Table~\ref{table:analogy}. 
we observed CWE only improved accuracy only for the family group.
The results are not surprising.
The words of family relations are compositional in Chinese, however capital and city names are usually not. 
We observed that GWE further improved CWE for words in the family group.
From Table~\ref{table:analogy}, we found that glyph features are helpful  when the characters can enhance word representations.
This is very reasonable because glyph features are fruitful representations of characters.
If character information does not play a role in learning word representations, character glyphs may not be useful.
The same phenomenon is observed in Table~\ref{table:wordsim}.

In our {\em job\&place}, we still observed that GWE improving CWE, however both CWE and GWE were slightly worse than CBOW. We also observed that Skipgram-based methods became worse than CBOW-based methods, while in all previous evaluation Skipgram-based methods are consistently better.

The results of RNN-Skipgram and RNN-GloVe are still poor.
We observe that the word representations learned from RNN  can  no longer be expressed by vector arithmetic. The reason is still under investigation.

\begin{CJK*}{UTF8}{bsmi}
\subsection{Case Study}
To further probe the effect of glyph features, we show the following word pairs in SimLex-999 whose calculated cosine similarities are higher based on GWE models than CWE.
The pairs may not look alike, but their components share related semantics. For example, in ``伶俐'' (clever), the component ``利''(sharp) is compositional to the meaning of ``俐''(acute), describing someone with a sharp mind. Other examples show the ability to associate semantics with radicals.

\begin{table}[h]
\centering
\scriptsize
\begin{tabular}{|l|c|c|}
\hline
{\bf Models} & {\begin{tabular}[c]{@{}c@{}}詞\\(word)\end{tabular} \& \begin{tabular}[c]{@{}c@{}}字典\\(dictionary)\end{tabular}} & {\begin{tabular}[c]{@{}c@{}}椅子\\(chair)\end{tabular} \& \begin{tabular}[c]{@{}c@{}}板凳\\(bench)\end{tabular}} \\\hline
CBOW 		 		& 0.2342 & 0.3469\\
\quad +CWE 		& 0.2918 & 0.3640\\
\quad +CWE+ctx\_Glyph	& \textbf{0.3361} & \textbf{0.3903}\\
\quad +CWE+tar\_Glyph	& 0.2857 & 0.3746\\
\hline
\hline
{\bf Models} & {\begin{tabular}[c]{@{}c@{}}鳥\\(bird)\end{tabular} \& \begin{tabular}[c]{@{}c@{}}火雞\\(turkey)\end{tabular}} & {\begin{tabular}[c]{@{}c@{}}聰明\\(smart)\end{tabular} \& \begin{tabular}[c]{@{}c@{}}伶俐\\(clever)\end{tabular}}\\\hline
CBOW & 0.2640 & 0.2634\\
\quad +CWE & 0.3064 & 0.2409\\
\quad +CWE+ctxG	& 0.3190 & 0.2710\\
\quad +CWE+ctxG	& \textbf{0.3422} & \textbf{0.2976}\\
\hline
\end{tabular}
\label{table:case}
\caption{Case study on word pairs in SimLex-999.}
\end{table}

\begin{table}[b]
\centering
\scriptsize
\begin{tabular}{|l|c|c|}
\hline
{\bf Models} & {\begin{tabular}[c]{@{}c@{}}山峰\\(mountain)\end{tabular} \& \begin{tabular}[c]{@{}c@{}}蜂蜜\\(honey)\end{tabular}} & {\begin{tabular}[c]{@{}c@{}}書桌\\(desk)\end{tabular} \& \begin{tabular}[c]{@{}c@{}}水果\\(fruit)\end{tabular}} \\\hline
CBOW 		 & 0.0581 & 0.0495\\
\quad +CWE 		& 0.0842 & 0.0719\\
\quad +CWE+ctxG	& 0.0736 & \textbf{0.0942}\\
\quad +CWE+tarG	& \textbf{0.1093} & 0.0733\\
\hline
\hline
{\bf Models} & {\begin{tabular}[c]{@{}c@{}}無趣\\(boring)\end{tabular} \& \begin{tabular}[c]{@{}c@{}}好笑\\(funny)\end{tabular}} & {\begin{tabular}[c]{@{}c@{}}胃\\(stomach)\end{tabular} \& \begin{tabular}[c]{@{}c@{}}腰\\(waist)\end{tabular}}\\\hline
CBOW & 0.3645 & 0.2388\\
\quad +CWE & 0.5351 & 0.2073\\
\quad +CWE+ctxG	& 0.5209 & 0.2500\\
\quad +CWE+ctxG	& \textbf{0.5426} & \textbf{0.2643}\\
\hline
\end{tabular}
\label{table:case_counter}
\caption{Counter examples to which GWE methods give higher similarity scores than CBOW or CWE.}
\end{table}

We also provide several counter-examples. Below are some word pairs which are not similar, however GWE methods produces higher similarity than CBOW or CWE. Take ``山峰'' (mountain) and ``蜂蜜'' (honey) as example. Since they share no common characters, the only thing in common is the component ``夆'', and we assume this to be the reason for the higher similarity. Also note that in the pair ``無趣'' (boring) and ``好笑'' (funny), the CWE similarity is also higher. We conclude that the character ``無'' (none) is not strong enough, so the character ``趣'' (fun) overrides the word ``無趣'' (boring), thus a higher score was mistakenly assigned.


\end{CJK*}

\section{Conclusions}
This work is a pioneer in enhancing Chinese word representations with character glyphs. The character glyph features are directly learned from the bitmaps of characters by convAE. 
We then proposed 2 methods in learning Chinese word representations: the first is to use character glyph features as enhancement; the other is to directly learn word representation from sequences of glyph features.
In experiments, we found the latter totally infeasible. Training word representations with RNN without word and character information is challenging. 
Nonetheless, the glyph features improved the character-enhanced Chinese word representations, especially on the word analogy task related to family. 

The results of exploiting character glyph features in word representation learning was ordinary. Perhaps the co-occurrence information in the corpus plays a bigger role than glyph features. 
Nonetheless, the idea to treat each Chinese character as image is innovative. As more character-level models\cite{zheng2013deep,Kim14,zhang2015character} are proposed in the NLP field, we believe glyph features could serve as an enhancement, and we will further examine the effect of glyph features on other tasks, such as word segmentation, POS tagging, dependency parsing, or downstream tasks such as text classification, or document retrieval.

\bibliography{emnlp2017}
\bibliographystyle{emnlp_natbib}

\end{document}